\documentclass[letterpaper, 10 pt, conference]{ieeeconf}  

\IEEEoverridecommandlockouts                              

\let\labelindent\relax
\usepackage{enumitem}
\usepackage{graphics} 
\usepackage{epsfig} 
\usepackage{mathptmx} 
\usepackage{times} 
\usepackage{amsmath} 
\usepackage{amssymb}  
\usepackage{graphicx}
\usepackage{tabularx}
\usepackage{multicol}
\usepackage{algorithm}
\usepackage{algorithmic}
\usepackage{multirow}
\usepackage{todonotes}%
\usepackage{lettrine}
\usepackage{subcaption}
\usepackage{wasysym}
\usepackage{bm}

\usepackage[font={scriptsize,it}]{caption} 
\usepackage[linewidth=1pt]{mdframed}
\usepackage{setspace}

\newcommand{\argmin}[1]{\underset{#1}{\operatorname{\mathbf{arg}}\,\operatorname{\mathbf{min}}}\;}
\newcommand{\x}{\mathbf{x}}

\newcommand{\ao}{\mathbf{u}}

\graphicspath{{pictures/}}
\usepackage{cite}
\usepackage[bookmarks=false]{hyperref}
\usepackage{leftidx}

\definecolor{lightlightgrey}{rgb}{0.9,0.9,0.9}
\definecolor{Red}{rgb}{1,0,0}
\definecolor{Blue}{rgb}{0,0,1}
\definecolor{Green}{rgb}{0,1,0}
\definecolor{magenta}{rgb}{1,0,.6}
\definecolor{lightblue}{rgb}{0,.5,1}
\definecolor{lightpurple}{rgb}{.6,.4,1}
\definecolor{gold}{rgb}{.6,.5,0}
\definecolor{orange}{rgb}{1,0.4,0}
\definecolor{hotpink}{rgb}{1,0,0.5}
\definecolor{newcolor2}{rgb}{.5,.3,.5}
\definecolor{newcolor}{rgb}{0,.3,1}
\definecolor{newcolor3}{rgb}{1,1,1}
\definecolor{darkgreen1}{rgb}{0, .35, 0}
\definecolor{darkgreen}{rgb}{0, .6, 0}
\definecolor{darkred}{rgb}{.75,0,0}

\xdefinecolor{olive}{cmyk}{0.64,0,0.95,0.4}
\xdefinecolor{purpleish}{cmyk}{0.75,0.75,0,0}

\xdefinecolor{red}{rgb}{1,0,0}

\usepackage{float}
\floatstyle{boxed}
\newfloat{math}{thp}{lop}
\floatname{math}{Equation}

\makeatother

\begin{document}

\title{Motion Planning for Multi-Mobile-Manipulator Payload Transport Systems}

\author{Rahul Tallamraju$^{1,4}$, Durgesh Haribhau Salunkhe$^2$, Sujit Rajappa$^3$, Aamir Ahmad$^4$, \\ Kamalakar Karlapalem$^1$, and Suril Vijaykumar Shah$^5$
\thanks{\hspace{-1em}\{rahul.tallamraju,aamir.ahmad\}@tuebingen.mpg.de, salunkhedurgesh@gmail.com,sujit.rajappa@uni-tuebingen.de, kamal@iiit.ac.in,surilshah@iitj.ac.in}
\thanks{\hspace{-1em}$^1$Agents and Applied Robotics Group, IIIT Hyderabad, India.}
\thanks{\hspace{-1em}$^2$Dibris Department, University of Genoa, Italy.}
\thanks{\hspace{-1em}$^3$Department of Computer Science, University of T\"ubingen, Germany.}
\thanks{\hspace{-1em}$^4$Max Planck Institute for Intelligent Systems, T\"ubingen, Germany.}
\thanks{\hspace{-1em}$^5$Mechanical Engineering Department, IIT Jodhpur, India.}  
}
\maketitle
\pagestyle{empty}
\begin{abstract}
In this paper, a kinematic motion planning algorithm for cooperative spatial payload manipulation is presented. A hierarchical approach is introduced to compute real-time collision-free motion plans for a formation of mobile manipulator robots. Initially, collision-free configurations of a deformable 2-D virtual bounding box are identified, over a planning horizon, to define a convex workspace for the entire system.  Then, 3-D payload configurations whose projections lie within the defined convex workspace are computed. Finally, a convex decentralized model-predictive controller is formulated to plan collision-free trajectories for the formation of mobile manipulators. This approach facilitates real-time motion planning for the system and is scalable in the number of robots. The algorithm is validated in simulated dynamic environments. Simulation video: {\footnotesize \url{https://youtu.be/9EKj7RwRs_4}}.
\end{abstract}


\section{Introduction} \label{sec:intro}
\begin{figure}[t] 
	\centering
	\includegraphics[scale=0.4]{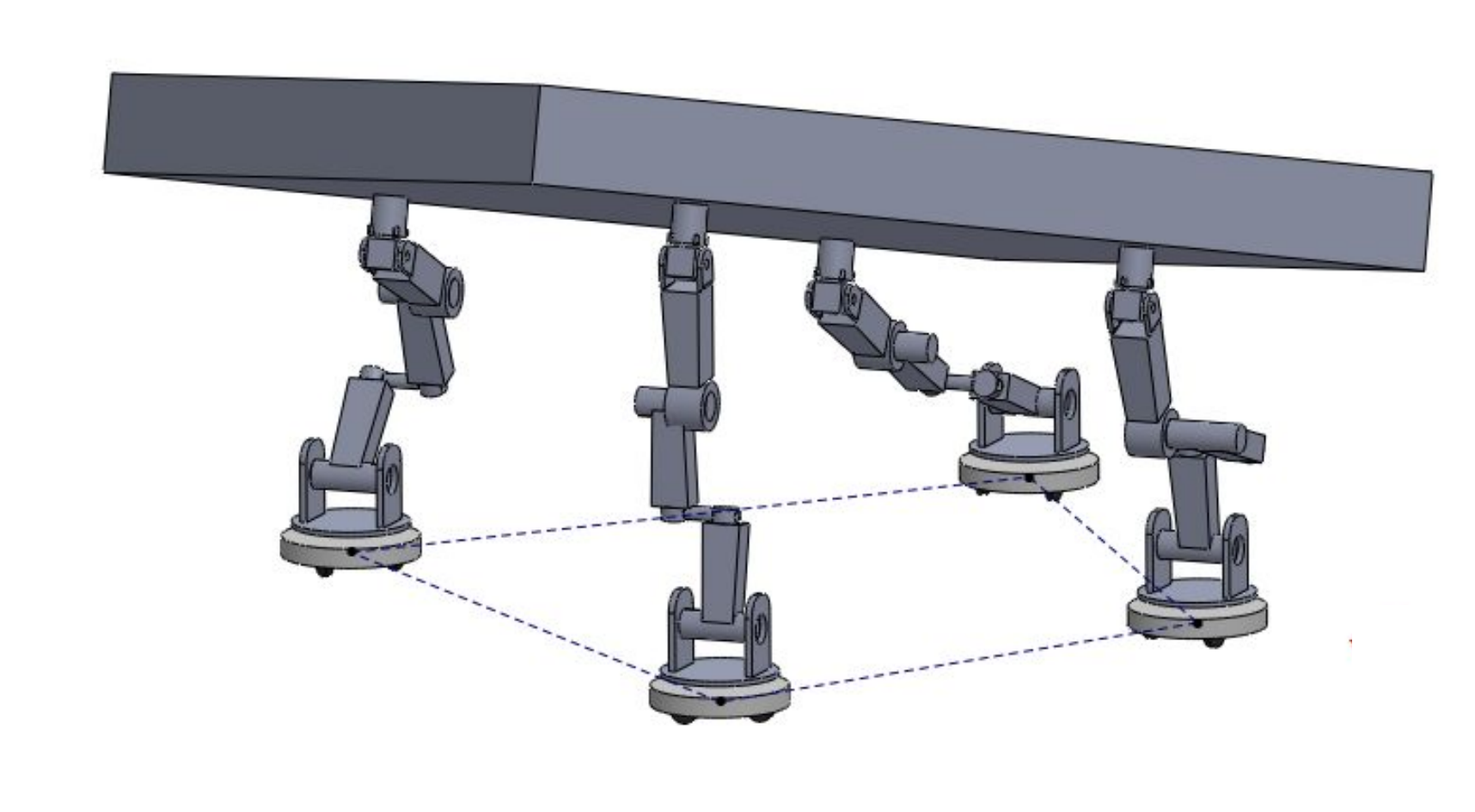} 
	\caption{A CAD illustration of the multi-mobile-manipulator payload transport system.}
	\label{f:cover}
\end{figure}

Coordination between robotic agents to collectively perform payload transportation tasks has recently piqued the interest of the robotics community \cite{alonso2016distributed,mellinger2013cooperative,jiao2017transportation}.
In this paper, we address the problem of local motion planning for cooperative payload manipulation in dynamic environments. 
Each robot in the system is associated with a 6-degrees-of-freedom (6-dof) manipulator and a holonomic mobile base. Each manipulator grasps a common payload as shown in Fig. \ref{f:cover}. The resulting system is a mobile parallel manipulator which is capable of non-planar payload manipulation. Planning real-time motion for such a setup in dynamic environments is challenging because of, (i) the high dimensional system configuration space, (ii) real-time environmental obstacles, and, (iii) inter-agent collision avoidance constraints which restrict the system to operate in a non-convex workspace.

In this work, we present three key contributions to address the above mentioned challenges.
(1) We propose a hierarchical motion planning algorithm which enables real-time non-planar mobile manipulation of a common payload in dynamic environments. 
(2) As part of this motion planning, we introduce a novel model-predictive controller (MPC) to identify real-time collision-free configurations of a deformable virtual bounding box. 
The planned bounding box configurations form a multi-scale convex workspace for the robots and the payload over the planning horizon.  
Subsequently, we identify feasible configurations of the payload which lie within the planned deformable virtual bounding boxes to guarantee environmental collision avoidance. 
(3) We then formulate a novel formation controller based on decentralized MPC to generate kinematically feasible, collision-free, trajectories for each robot. The manipulator configuration associated with each robot is determined using inverse kinematics between the payload grasp point and the position of the mobile base. 

We validate the efficacy of our hierarchical motion planner by simulating the entire mobile parallel manipulator setup in different environments consisting of both static and dynamic obstacles. The robots are tasked to safely manipulate and navigate the payload while tracking a moving target. 


\section{Related Work} \label{sec:related_work}
In this section, the works pertinent to multi-robot payload transport systems are reviewed in detail. \\
\textit{Formation Frameworks:} Multi-robot payload transportation constraints mobile robots to navigate in a formation. Leader-follower \cite{desai2001modeling}, behavior-based \cite{balch1998behavior}, virtual leader-follower \cite{bhatt2009formation} and virtual structure \cite{lewis1997high} are some of the formation frameworks that have been proposed in literature.   
We employ the virtual leader-follower framework in our work to derive decentralized motion plans. 

\textit{System Modeling and Control: } Virtual linkage models \cite{khatib1996coordination} were used to identify multi-grasp forces for cooperative manipulation and control.
Flexible object transportation with static collision avoidance using non-holonomic mobile manipulators was explored in \cite{tanner2003nonholonomic}.
In \cite{bhatt2005screw}, a screw theoretic framework was developed for planar payload mobile manipulation using multiple robots mounted with passive 2-dof manipulators. 
Adaptive controllers \cite{fang2008adaptive} were studied for dealing with unknown payload inertia parameters, external disturbances and tracking a known trajectory using multiple robots. 
The above works do not consider non-planar rigid payload manipulation coupled with spatiotemporal motion planning constraints.

\textit{Constrained Optimization: } In \cite{bhatt2009formation},  differential geometry based formation optimization was used to compute energy-efficient robot trajectories.  Semi-definite \cite{derenick2010semidefinite} programming was leveraged to plan motion for robot formations, with constraints on shape templates, network-connectivity, and obstacle avoidance. Trajectory optimization \cite{morgan2016swarm} with linearized dynamics was solved as a sequential convex program to assign tasks and plan motion of robot swarms. Offline trajectory optimization was established for planar payload transportation using multiple passive 1-dof non-holonomic mobile manipulators \cite{abbaspour2015optimal}. An integer program \cite{jiao2017transportation} was proposed to geometrically plan motion for multi-mobile manipulator, non-planar payload transportation. In \cite{alonso2015multi,alonso2016distributed} multi-robot planar mobile manipulation of a payload  was achieved using non-linear optimization , which is solved as a sequential convex program. 
An extension to \cite{alonso2015multi} was probed to holonomically manipulate deformable payloads \cite{alonso2015local}. \\
In contrast to the above works, each mobile manipulator in our system has the freedom of having significant relative motion between its end-effector grasp point and the mobile base. This freedom grants us the ability to plan kinematically feasible and energy optimal robot trajectories. We formulate a scalable motion planning framework which enables a formation of robots to navigate through dynamic environments while manipulating a payload in 3-D space.\\
Sec. \ref{sec:prelims} briefs on the mathematical notations and the hierarchical motion planning algorithm, Sec. \ref{sec:manipulation} analyzes the mobile parallel manipulator, Sec. \ref{sec:poa}  discusses the proposed hierarchical motion planning approach in detail, Sec. \ref{sec:results} elaborates on simulation results. We conclude in Sec. \ref{sec:conclusions} and discuss future work directions.    


\section{Overview} \label{sec:prelims}
\subsection{Preliminaries}
\begin{figure}[t] 
	\centering
	\begin{subfigure}{0.4\columnwidth}
		\centering
		\includegraphics[scale = 0.31,trim={0cm, 2cm, 0cm, 0cm}]{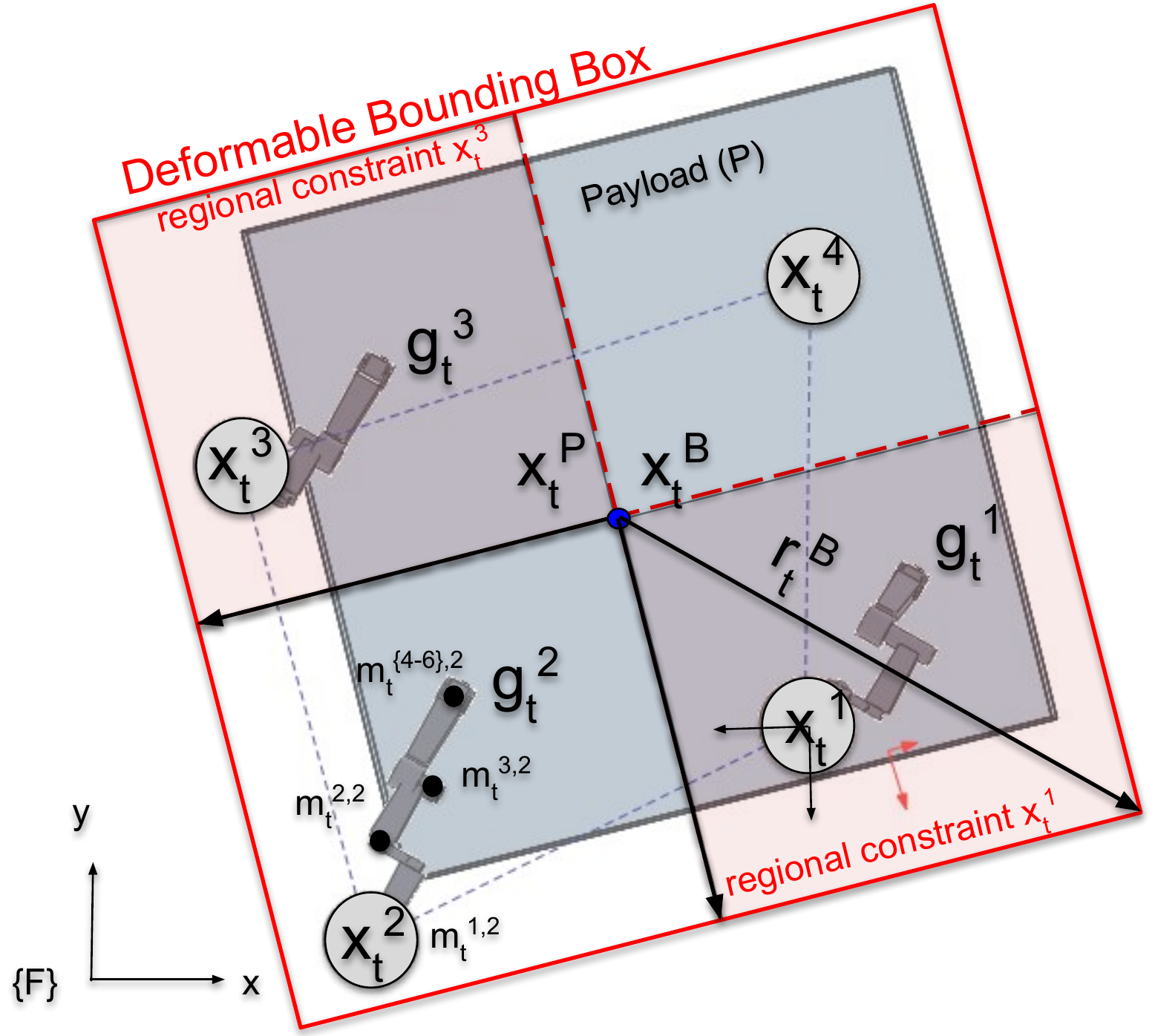} 
		\vspace{1em}
		\caption{}		
		\label{f:prelim}
	\end{subfigure}~\hspace{1.1em}
	\begin{subfigure}{0.6\columnwidth}
		\centering
		\includegraphics[scale = 0.48]{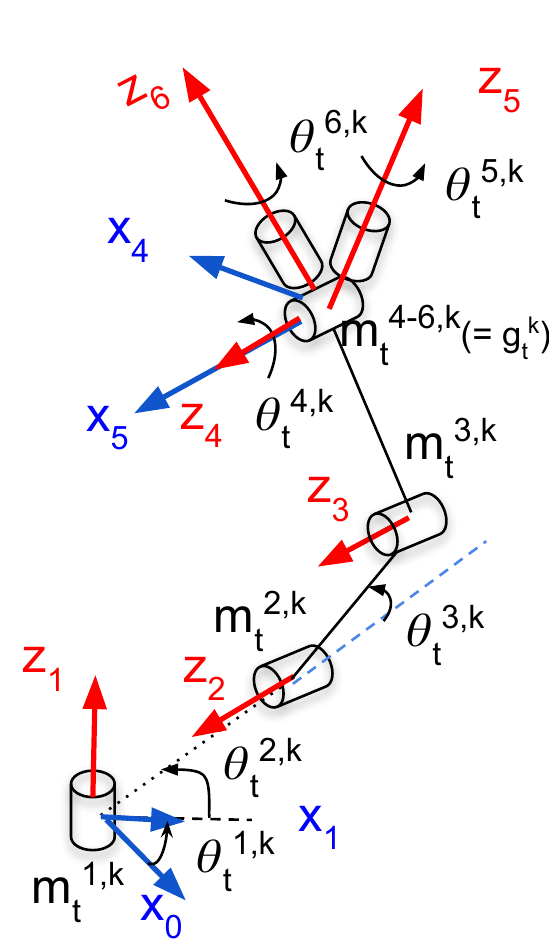} 
		\caption{}
		\label{f:manip_schem}
	\end{subfigure}
	\caption{(a) Notations used in this work are overlaid on an illustration of a four robot multi-mobile-manipulator system. (b) Manipulator configuration schematic.}
	\label{f:prelims}
\end{figure}

Let there be $K$ mobile manipulator robots transporting a payload $P$ while tracking a moving target $T$. The position of $T$ in the world frame at time $t$ is denoted by $\x_t^T \in \mathbb{R}^2$. Typically the moving target is a person guiding the system of robots globally. The configuration $\xi_t^{k}$ of each robot $k\in[1\dots K]$ is defined by the pose (position $\x_t^{k}$ and yaw $\psi_t^{k}$) of the mobile base and the manipulator joint angles $\theta_t^{i,k}, \forall i \in [1,6]$, $\xi_t^{k}=[\x_t^{k} ~ \psi_t^{k} ~ \theta_t^{1,k}..\theta_t^{6,k}]\in\mathbb{R}^9$. The position of the manipulator joints in the task space are denoted as $\mathbf{m}_t^{i,k} \in \mathbb{R}^3, ~ \forall i \in [1,6]$. Each robot $k$ grasps $P$ rigidly at $\mathbf{g}_t^k \in \mathbb{R}^3$ $(=\mathbf{m}_t^{6,k})$, which is pre-defined and lies on the surface of $P$. The system of robots is enclosed by a 2-D virtual bounding-box $B$ whose configuration is denoted by $\xi_t^{B} = [ \x_t^{B}, ~ \psi_t^{B}, ~ r_t^B]$ where $\x_t^{B}$ denotes the 2D position of its centroid, $\psi_t^{B}$ its yaw and $r_t^B$ the length of its half-diagonal (representing scale). The payload's geometric center $\x_t^P\in\mathbb{R}^3$ is at a constant height offset $h^P$ above $\x_t^B$. The yaw $\psi_t^P$ is always aligned with $\psi_t^B$, s.t. $\psi_t^P\cong\psi_t^B$. Fig. \ref{f:prelims} provides an overview of the notations described above on an illustration of a four robot multi-mobile manipulator system. Regional constraints of each mobile manipulator are also visualized in Fig. \ref{f:prelim}, definitions of which are further discussed in Sec. \ref{sec:dmpc}.

\begin{figure*}[t] 
	\centering
	\includegraphics[width=\linewidth]{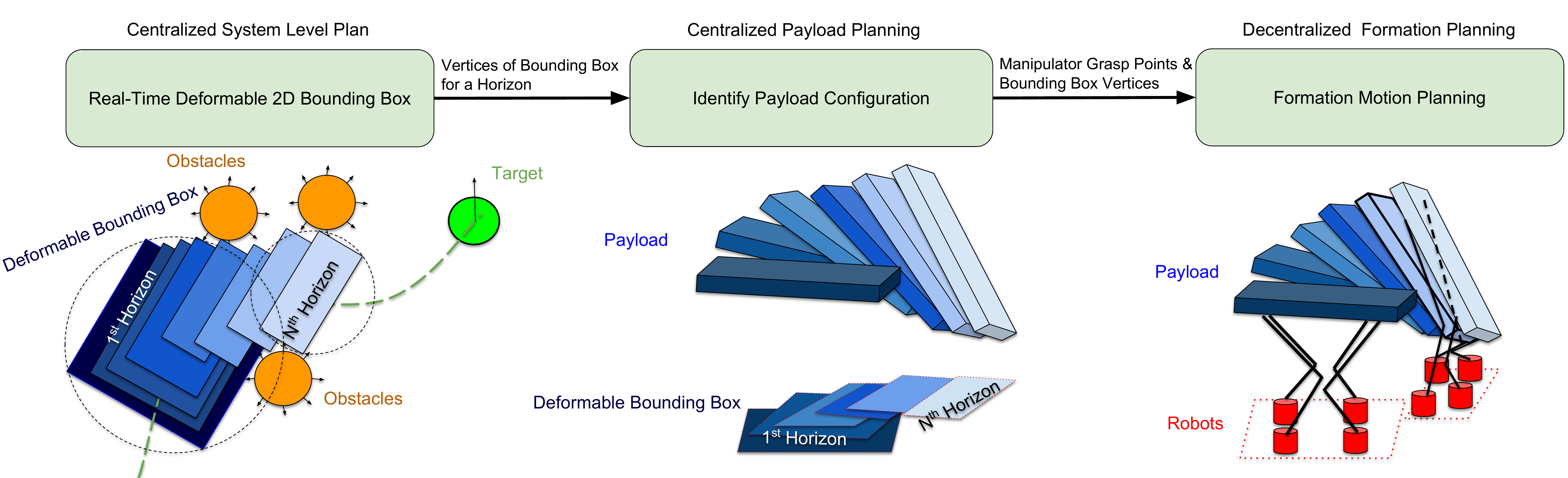} 
	\caption{Overview of the proposed hierarchical motion planning approach. First, the deformable bounding box configurations for a fixed time horizon are determined. Second, payload configurations which lie within the bounding box are computed. Finally, robot trajectories are derived to facilitate cooperative manipulation of the payload.}
	\label{f:plan}
\end{figure*}

\subsection{Motion Planning Overview} 
The operational environment consists of $M$ static obstacles and $N$ non-cooperative dynamic obstacles $(S_1,..S_M,D_1,..D_N)$. The objective of the system of robots is to ensure that the centroid of the system bounding-box $\x_t^{B}$ reaches a desired destination position $\x_t^{B_d}$ in the vicinity of the target position $\x_t^{T}$. Our work is motivated by the application of simultaneous target tracking (\hspace{-0.01cm}\cite{price2018deep,tallamraju2019active}) and payload transportation. The key requirements in our target tracking scenario are, (i) to not lose track of the target, and, (ii) to ensure that the payload and the formation of robots avoid all the obstacles in their vicinity. To address both these objectives in an integrated approach, we formulate the hierarchical motion planning algorithm, as detailed in Fig. \ref{f:plan}. The main steps are,
\begin{enumerate}
	\item compute $\x_t^{B_d}$ in the direction of the target position,
	\item avoid environmental obstacles with the deformable virtual bounding box using a novel MPC (section \ref{sec:dvb}) for navigation and deformation,
	\item rotate the payload to ensure that its projection lies within the area occupied by $B$ over the horizon (section \ref{sec:pmp}),
	\item solve a novel decentralized MPC (section \ref{sec:dmpc}) to compute efficient and collision-free mobile base trajectories,
	\item identify manipulator configurations based on known mobile base positions $\x_t^k$ and the derived end-effector grasp positions $\+g_t^k$ from the payload configuration.
\end{enumerate}

\section{System Description: Mobile Parallel Manipulator} \label{sec:manipulation}
Each robot in the system consists of a holonomic 3-dof mobile base and a 6-dof manipulator. Based on the definition of $\xi_t^k$, each robot has a configuration space $\mathtt{C}^{robot}\in\mathbb{R}^9$. Since there are $K$ robots in the system, $\mathtt{C}^{system} \in  \mathbb{R}^{9 \times K}$. The system motion is planned for a fixed prediction horizon $H$ to compute trajectories for ensuring inter-robot and environmental collision avoidance (M static obstacles and N dynamic obstacles). Therefore, at every time instant the system has to identify configurations in $\mathtt{C}^{plan} \in \mathbb{R}^{9\times K \times H}$, while also being subjected to non-convex collision avoidance and kinematic constraints. For real-time motion planning, this is computationally expensive and is not scalable in the number of robots. We, therefore, decentralize the computation of robot configurations. This restricts the search space to $\mathbb{R}^{9\times H}$, albeit by increasing the communication complexity. If both the mobile base control inputs and the angular velocity of the payload (non-planar) orientation are minimized, there is an implicit energy minimization of the manipulator's joint velocities. Therefore, we hierarchically optimize the payload orientation, and subsequently, the position of each mobile base in $\mathtt{C}^{traj}\in\mathbb{R}^{3\times H}(\subset \mathtt{C}^{plan})$ to obtain sub-optimal but feasible systemic configurations.




\noindent \textbf{Remark 1} - \textit{We observe that, by using any  number ($K$) of  revolute 6-dof manipulators (in non-singular configurations) to rigidly grasp and manipulate a common payload, the system always has at least 6-dof. This key observation enables us to hierarchically plan the motion of the payload, mobile bases and the manipulators. We analyze this remark and its consequences below.} 

Analysing our system of robots as a static parallel manipulator which is rigidly affixed to the ground, the dof of the constrained system in accordance with Gr\"ubler's formula is,
\begin{equation}
dof_s = \lambda(l-1-j) + \sum\limits_{j}dof_i~,
\label{eq:Grubler}
\end{equation}
where, $\lambda = 6$ (for spatial systems), $l$ is the total number of links, including all the fixed links, $j$ is the total number of joints in the system, and
$dof_i$ and $dof_s$ denote the degrees of freedom of each joint and the system respectively. In our spatial system, each manipulator $k$ has five links and six revolute joints $(dof_k=6)$. Treating the robot bases and ground plane as a single link and the payload as another link, Eq. (\ref{eq:Grubler}) for the static parallel manipulator is computed as,
\begin{equation}
dof_s = 6((5K + 2)-1-6K) + 6K~,
\end{equation}    
where, $K$ is the number of manipulators grasping the payload. \textit{Notice that the dof of the system ($dof_s$) is six for any $K$}. This result has the following important consequences: \\
(a) \textit{Scalability:} The mobility of the parallel manipulator is not affected by the number of robots used. Therefore, the system is kinematically scalable. \\
(b) \textit{Payload Mobility:} Since the parallel manipulator has 6-dof, the payload has complete mobility in 3-D space. This is limited only by the manipulator link lengths.\\
(c) \textit{Robot Mobility:} As $dof_s=dof_k$, there exists a natural partition in the task space of the bases and manipulators. Mobile bases (3-dof) extend the degrees of freedom of each manipulator. This mobility introduces significant relative motion for the bases w.r.t. to the manipulator grasp points. This is limited only by manipulator link lengths and spatiotemporal constraints (e.g., obstacle avoidance). \\
(d) \textit{Decentralization:} Each robot can independently compute feasible and efficient motion plans for itself. \\

\noindent \textbf{Remark 2} - \textit{For the manipulated common payload to have at least 6-dof mobility, each manipulator $k\in [1\dots K]$ should have at least 6-dof. This remark is conditioned on each joint being revolute and the manipulator being in a non-singular configuration.}\\
If each manipulator $k$ has $j_k$ revolute joints, $l_k$ links and $dof_k$ degree-of-freedom . For a system with $K$ robots to have atleast 6-dof, Eq. (\ref{eq:Grubler}) is evaluated as,
\begin{eqnarray}
6((l_kK + 2)-1-j_kK) + K(dof_k) & \geq & 6 \nonumber \\
6(l_k-j_k) + dof_k & \geq &0~.
\end{eqnarray}
In our case $(l_k-j_k) = -1$, which implies $dof_k\geq6$. Therefore each manipulator should have at least 6 revolute joints to ensure that the payload has 6-dof. 


\section{Hierarchical Motion Planning} \label{sec:poa}
In this section, we describe our proposed hierarchical motion planner in detail.
Sec. \ref{sec:dvb} describes a convex optimization formulation to plan the motion of a deformable virtual bounding box (DVB) through static and dynamic obstacles. High-level motion guidance for the DVB is achieved by tracking a desired moving target. Obstacle avoidance is enforced by embedding artificial repulsive potential fields into the optimization, which aid in deforming and navigating the DVB. Sec. \ref{sec:pmp} presents an optimization program to identify non-planar configurations of the payload so as to ensure that its projections lie within the convex workspace defined by DVB. In Sec. \ref{sec:dmpc} feasible, collision-free robot trajectories are computed using a convex decentralized MPC for the mobile bases.
 
\subsection{Deformable Virtual Bounding Box} \label{sec:dvb}
To identify collision-free goal-directed motion plans for the DVB, we formulate an MPC. The objective of the MPC is to ensure that the centroid $\x_t^B$  reaches a desired position $\x_t^{B_d}$ efficiently. The MPC is constrained by obstacle avoidance, linear translational dynamics, position limits, and control saturation bounds. 
\subsubsection{Objective}
A desired position $\x_t^{B_d}$ of the bounding box in the vicinity of the target $\x_{t}^{T}$ is defined as,
\begin{equation} \label{eq:despose}
\x_t^{B_d} = \x_{t}^T - \left[d_{des}cos(\psi_t^{T,B})~~ d_{des}sin(\psi_t^{T,B})\right]~,
\end{equation}
where, $\psi_t^{T,B}$ is the angle of $x_t^B$ about the target position $x_t^T$ and $d_{des}$ is a desired distance to the target. 
The cost of the MPC is given as,
\begin{align} \label{eq:J_MPC}
&J_{\mathrm{MPC}} =  \sum_{n=0}^{H} (\ao_t^{B}(n) \boldmath{\Omega}_{u} \ao_t^{B}(n)^\top  + \nonumber \\ &(\x_t^{B}(n+1)  - \x_t^{B_d}) \boldmath{\Omega}_{x} (\x_t^{B}(n+1)  - \x_t^{B_d})^\top), 
\end{align}
where, $\ao_t^B(n)\in \mathbb{R}^2$ is the linear translational velocity control input at discrete horizon step $n$. $\boldmath{\Omega}_{u}, \boldmath{\Omega}_{x}$ are diagonal positive semi-definite weight matrices for the control and state costs respectively. Eq. (\ref{eq:J_MPC}) minimizes the control input and the distance between $\x_t^B$ and $\x_t^{B_d}$ over a fixed time horizon $H$.

\subsubsection{Environmental Obstacle Avoidance}
We compute artificial repulsive potential field  vectors for a planning horizon and embed them as external control inputs in the MPC dynamics constraint. This operation preserves the convexity of optimization. Incorporating external control inputs to avoid collisions within an MPC framework was presented in our previous work \cite{rahult}. 
The repulsive potential field magnitude w.r.t the $i^{th}$ obstacle, is given as,

{\small \begin{equation}\label{eq:cotforce}
	F^{i}(d) =  
	\begin{cases}
	F_{max}\ & \text{if} ~~ d < d_{min} \\
	\frac{\pi}{2} \big(\frac{cot(z)+z -\frac{\pi}{2}}{d_{max}-d_{min}}\big)& \text{if} ~~ d_{min} \leq d \leq d_{max} \\
	0, & \text{if} ~~ d > d_{max}
	\end{cases}\;.
	\end{equation}}
Here, $z = \frac{\pi}{2} \big(\frac{d-d_{min}}{d_{max}-d_{min}}\big)$, argument $d$ is a distance metric between $\x_t^B$ and obstacle $i$. Note that, $	F^{i}(d)$ varies hyperbolically w.r.t $d$. $d_{max}$ and $d_{min}$ are distances defining the region of influence of the potential field and the distance at which the potential field value tends to infinity respectively. In practice, the potential field at $d_{min}$ is clamped to a positive value $F_{max}\geq max(\|\ao_t^{B}\|)$ to ensure obstacle avoidance. \\
\textit{Dynamic Obstacle Avoidance}:
The external control input $\mathbf{f}_{dyn}(n)$ due to the influence of $N$ dynamic obstacles is,

{\small \begin{equation}\label{rep_dyn}
	\mathbf{f}_{dyn}(n) = 
	\sum_i^NF^{D_i}(d_{dyn}(n))\;\mathbf{\beta}^{D_i}(n)~,
	\end{equation}
}
where, argument {\small $d_{dyn}(n) = \|\x_t^B(n) - \x_t^{D_i}(n) \|_2, \forall n \in [1,H]$}, is computed using both the prediction horizon motion plan for the bounding box and the dynamic obstacle $D_i$.
The external control input acts along the direction {\small $\mathbf{\beta}^{D_i}(n) = \frac{\x_{t}^{B}(n)-\x_t^{D_i}(n)}{\|\x_{t}^{B}(n)-\x_t^{D_i}(n)\|_2}$}, which is a unit  vector pointing in the direction away from $D_i$'s horizon motion plan\footnote{The state evolution model and velocity of $D_i$ are assumed to be known.}. $d_{min}$ is the length of the half-diagonal $r_t^B(n)$ of B (see Fig. \ref{f:multi_scale_mpc})  and $d_{max}=d_{min}+\delta, \delta>0$. \\
\textit{Static Obstacle Avoidance}:
The total external control input due to the $M$ static obstacles is given by,

{\small \begin{equation}\label{rep_sta}
	\mathbf{f}_{sta}(n) = 
	\sum_i^MF^{S_i}(d_{sta}(n))\;\mathbf{\beta}^{S_i}(n)~,
	\end{equation}
}
where, {\small $d_{sta}(n)= \|\x_t^B(n) - \x_t^{S_i}\|_2$} and {\small $\mathbf{\beta}^{S_i}(n) = \frac{\x_{t}^{B}(n)-\x_t^{S_i}}{\|\x_{t}^{B}(n)-\x_t^{S_i}\|_2}$} . 
The tracked target is also considered as a static obstacle.
\subsubsection{Handling Field Local Minima}
\begin{figure}[t]
	\centering
	\includegraphics[width=\columnwidth,trim={0cm, 0cm, 0cm, 1cm}]{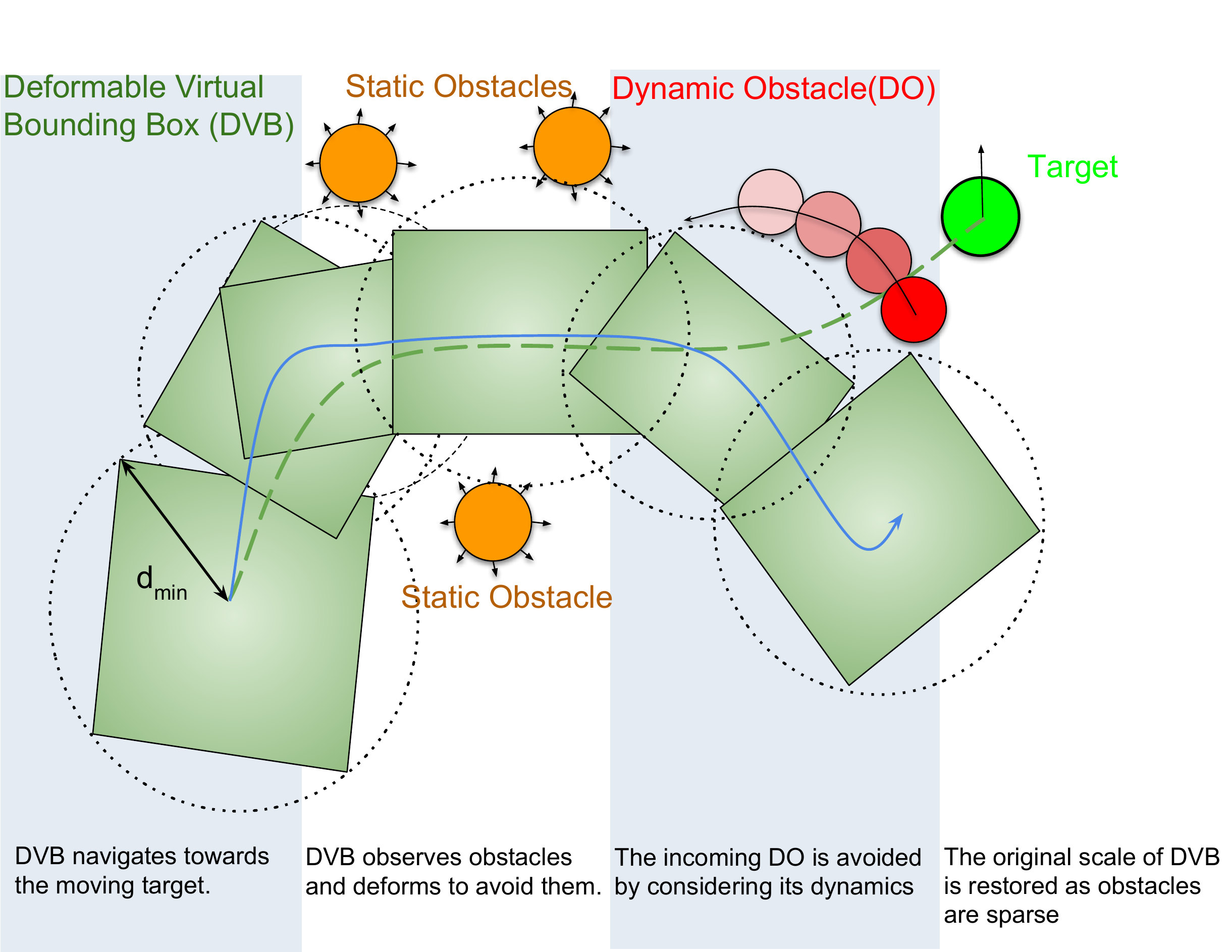}
	\caption{DVB deforms and navigates through obstacles while tracking the moving target.}
	\label{f:multi_scale_mpc}
\end{figure} 
In cluttered static environments the DVB could stagnate in space due to field local minima. We propose the following approaches to resolve this.\\
\textit{Approach Angle Force:} Repulsive potential fields are used to deviate the DVB away from obstacles which lie along the direction of approach to the target. From (\ref{eq:cotforce}), we utilize $F^i(d)$ to compute the repulsive force $\+f_{ang}(n)$ as a function of absolute angular difference $d_{ang}=|\psi_t^{T,B}-\psi_t^{T,S_i}|$.  $\psi_t^{T,B}(n)$ is the angle of $B$ w.r.t to $T$, $\psi_t^{T,S_i}(n)$ is the angle of $S_i$ w.r.t $T$. We term this external control input as approach angle force (introduced in \cite{rahult}). The total obstacle avoidance external control input on $B$ is given by,
\begin{equation}
\mathbf{f}^{B}_{rep}(n)=\mathbf{f}_{dyn}(n)+\mathbf{f}_{sta}(n)+\textbf{f}_{ang}(n)
\end{equation}

To embed this control input into the MPC dynamics we clamp the magnitude of $\mathbf{f}^{B}_{rep}(n)$ to $F_{max}$ (defined in (\ref{eq:cotforce})).
{\small \begin{equation} \label{clamping}
	\mathbf{f}_{t}^{B}(n)=
	\begin{cases}
	\mathbf{f}^{B}_{rep}(n)  & \text{if} ~~ \|\mathbf{f}^{B}_{rep}(n)\| < F_{max} \\
	F_{max}\frac{\mathbf{f}^{B}_{rep}(n)}{\|\mathbf{f}^{B}_{rep}(n)\|} & \text{if} ~~  \|\mathbf{f}^{B}_{rep}(n)\| \geq F_{max}         
	\end{cases}
	\end{equation}}
Note that in practice, we observed that the external input $\+f_t^B(n)$ could cause high-frequency oscillations in the horizon motion plans if $B$ is very close to obstacles. To reduce these oscillations, we add a fraction of external control input from the previous time step $\kappa \+f_{t-1}^B(n), 0<\kappa<1$. The total external input which is clamped at $F_{max}$ is now given by $\+f_{t}^{B}(n)=\+f_{t}^{B}(n)+\kappa \+f_{t-1}^{B}(n)$. This operation recursively reduces the effect of obstacles on $B$ over multiple time steps and therefore ensures a smooth change in external control input. \\
\textit{Bounding-Box Deformations:} The DVB can be deformed in size by regulating $r_t^B$ over the horizon. Reducing $r_t^B(n)$, instantaneously reduces the region of influence of the potential field. This enables $B$ to navigate through tight spaces. Fig. \ref{f:multi_scale_mpc} highlights the $d_{min}$ of the potential field (dotted-circle) as the DVB navigates between obstacles. The deformation in $r_t^B$ is applied over the MPC horizon, and the rate of this deformation is proportional to the total repulsive potential field value $\mathbf{f}_{t}^{B}(n)$. In our work, the DVB is deformed only along its width. For ease of navigation of the mobile manipulators, it is important to restore $B$ to its original size in regions where obstacles are sparse. Therefore, $B$ is also associated with a small expansion potential field $f_t^e(n)$, which is directed towards increasing its width. 
This DVB width $w^B_t(n)$ and scale $r_t^B(n)$ over a horizon are computed as follows.
{ \begin{eqnarray}
	w^B_t(n+1) &=& w^B_t(n) - k_{s}\|\+f^{B}_{t}(n)\|_2 + k_e f^{e}_{t}(n)\\
	r_t^{B}(n+1) &=& 0.5\|[l^B,w^B_t(n+1)]\|_2
	\end{eqnarray}}
where, $l^B, w^B_t(n)$, refer to the length and variable width of the DVB respectively. $k_s,k_e$ are positive constants with $k_s > k_e$. The region of influence of $f_t^e(n)$ is given by $d_{min}= min(r^{B}_t)$ and $d_{max}= max(r^{B}_t)$ which are the minimum and maximum allowed deformations respectively. 

\subsubsection{MPC Formulation}
\small
\begin{equation} \label{eq:MPCopt}
\x_t^{B*}(1)\dots\x_t^{B*}(H+1),\ao_t^{B*}(0)\dots\ao_t^{B*}(H) = \argmin{\ao_t^{B}(0)\dots\ao_t^{B}(H)} (J_{MPC})
\end{equation} 
subject to, 
\begin{align}
& \x_t^{B}(n+1)^\top  =  \mathbf{A} \x_t^{B}(n)^\top  + \mathbf{B}(\ao_t^{B}(n)+\mathbf{f}_{t}^{B}(n))^\top, \label{eq:LTI}  \\
& \ao^B_\mathrm{min} \leq ~ \ao_t^{B}(n) \leq \ao^B_\mathrm{max}, \\
& \x^B_\mathrm{min} \leq ~ \x_t^{B}(n+1) \leq ~ \x^B_\mathrm{max}. \label{eq:last_MPCopt}
\end{align}
\normalsize
The objective of the MPC is given by (\ref{eq:J_MPC}). The motion planning adheres to the following constraints.
\begin{enumerate}
	\item LTI dynamics of $\x_t^B$ given by (\ref{eq:LTI}),
	\item collision avoidance constraints incorporated as an external control input $\mathbf{f}_t^B \in \mathbb{R}^2$,
	\item position and control saturation bounds on $\x_t^B$, $\ao_t^B$.
\end{enumerate}
Dynamics ($\mathbf{A} \in \mathbb{R}^{2\times2}$) and control transfer ($\mathbf{B} \in \mathbb{R}^{2\times2}$) matrices are given by $\+A = \+I_{2\times2}$ and $\+B = \Delta t  \+I_{2\times2}$ where, $ \+I_{2\times2}$ is an identity matrix and $\Delta t$ is the sampling time. The quadratic program computes optimal control inputs $\left[\ao_t^{B*}(0) \cdots \ao_t^{B*}(H) \right]$ and trajectory $\left[ \x_t^{B*}(1) \cdots \x_t^{B*}(H+1) \right]$ towards the $\x_t^{B_d}$. The  scale $r_t^B(1)$ is used as the initial DVB state for the next optimization iteration at $t+1$. The position $\x_t^B$ and yaw $\psi_t^B$ are then controlled using a proportional controller with $\x_t^{B*}(1)$ and $\psi_t^{T,B}$ (defined in (\ref{eq:despose})) as desired position and yaw.

\subsection{Payload Motion Planning} \label{sec:pmp}
The DVB described in the previous section forms a multi-scale convex workspace for the system of robots and the payload over the MPC horizon. The primary goal of payload motion planning is to roll the payload $P$ to ensure that its projection $\hat{P}$ lies within the planned DVBs. This guarantees environmental collision avoidance for $P$. Accordingly,  a non-linear optimization program with the objectives of minimizing payload roll $\phi_t$ and angular velocity  $\omega_t$ over a planning horizon $H_p \leq H$ is formulated. The optimization is constrained to ensure $\hat{P}$ lies within the computed DVB for a horizon. It is important to note that the limit on $w_t^B$ is defined in accordance with the maximum roll $\phi_{max}$ of the payload. This guarantees the existence of a feasible solution to $\phi_t$. 
The constrained optimization is formulated as,
{\begin{align} 
&\phi^*_t(n), ~ \omega^*_t(n) = \argmin{\omega_t(0) \dots \omega_t(H_p)}  \sum_{n=0}^{H_p} \text{w}_1 \phi_t(n+1)^2 + \text{w}_2 \omega_t(n)^2
\nonumber \end{align} }
subject to, 
{\small \begin{align}
& \mathbf{Reg}_t^B(n+1) \hat{T}_P^F (\phi_t(n+1),\x_t^P(n+1),\mathbf{v}_i)^\top \leq \mathbf{b}_t^B(n+1), \forall i,  \label{eq:regcon} \\
& \phi_t(n+1) = \phi_t(n) + \omega_t(n) \Delta t, \label{roll_dynamics}\\
& \phi_{min} \leq \phi_t(n+1) \leq \phi_{max}, \\
& \omega_{min} \leq \omega_t(n) \leq \omega_{max}, \label{eq:pmp}
\end{align}}
where, $\text{w}_1,\text{w}_2$ are constant positive weights on the objectives. Here, (\ref{eq:regcon}) defines linear polygonal regional constraints for the vertices of the payload $\forall n$. For each horizon step, $\mathbf{Reg}_t^B(n+1) \in \mathbb{R}^{4\times2}$ and $\mathbf{b}_t^B(n+1)\in \mathbb{R}^{4}$ represent the planned DVBs. $T_P^F$ is a homogeneous transformation (in SE(3)) of a point $v_i$ (defined in the payload local frame) to the world frame ${F}$, for a roll of $\phi_t(n+1)$ and a translation of $\+x_t^P$.  $\hat{T}_P^F$ computes the projection of the transformed point onto the DVB. Here, $v_i$ are points which are defined in the payload local frame, to parameterize the payload. In our work, a cuboidal payload is used and parameterized using its vertices $v_i, i\in [1,8]$. The extension to non-cuboidal payloads is straight-forward as long as its 3-D convex hull vertices are known. The constraint in (\ref{roll_dynamics}) controls the time evolution of $\phi_t$, subject to limits on $\phi_t$ and $\omega_t$. The defined optimization minimizes the weighted sum of squares of $\phi_t$ and $\omega_t$ to ensure that payload rolls smoothly while staying within the planned DVBs (see Fig. \ref{f:plan}). The optimal solution at the first horizon step (i.e. $\phi^*_t(1)$) is used as the roll for time instant $t$. 

\subsection{Decentralized Formation Motion Planning}\label{sec:dmpc}
In this section we first introduce the inverse kinematics of the 6-dof manipulator used in this work and subsequently discuss the decentralized formation motion planning.
\subsubsection{6-dof Manipulator}\label{sec:invkin}
The 6-revolute joints are visualized in Fig. \ref{f:DHparams}. The joint configuration is similar to industrial 6-dof manipulators like the KR3 Agilus\footnote{\url{http://www.kuka.com/}}  or the UR-3 manipulator\footnote{\url{http://www.universal-robots.com/}}.
The inverse kinematics (ikin) of the manipulator for a known grasp point $\mathbf{g}^k_t$ and mobile base position $\x^k_t$ is analytically determined as follows. 
\begin{itemize}
\item $\theta_t^{1,k}$, is computed using the orientation of the vector ($\mathbf{g}^k_t-\x^k_t$). 
\item $\theta_t^{2,k}, \theta_t^{3,k}$ form a planar two-link manipulator in the plane defined by $\theta_t^{1,k}$. $\theta_t^{3,k} = \cos^{-1}\frac{(R_z(\theta_t^{1,k})g^k_t)_x^2+(R_z(\theta_t^{1,k})\+g^k_t)_z^2-l^2_2-l^2_3}{2l_2l_3}$. $l_2=\|m_t^{2,k}-m_t^{3,k}\|_2$, $l_3=\|m_t^{3,k}-m_t^{4,k}\|_2$ are link lengths and $R_z(\theta_t^{1,k})$ is a three dimensional rotation matrix about the first joint axis.  
$\theta_t^{2,k} = \tan^{-1}\frac{(R_z(\theta_t^{1,k})\+g^k_t)_z}{(R_z(\theta_t^{1,k})\+g^k_t)_x} - \tan^{-1}\frac{l_3 \sin{\theta_t^{3,k}}}{l_2 + l_3 cos\theta_t^{3,k}}$.
\item $\theta_t^{4-6,k}$ form a wrist configuration without any offset, whose values are determined based on the payload orientation. $\theta_t^{4,k}, \theta_t^{5,k}$ are solved as spherical angles of the vector $(\x_t^P-\mathbf{g}_t^k)$ and $\theta_t^{6,k}$ is along this vector, which is solved using the plane formed by the payload. 
\end{itemize}

\subsubsection{Inter-Base Collision Avoidance}
The planned DVB for a horizon using (\ref{eq:MPCopt}-\ref{eq:last_MPCopt}) form a collision-free convex hull for the robot formation. We define an operational regional constraint within $B$ for each mobile base as shown in Fig. \ref{f:prelim}. This region is undilated by the radius\footnote{A mobile base is defined as a cylinder of a certain radius and height} of the robot to guarantee inter-mobile-base collision avoidance (visualized in Fig. \ref{f:avoidance}). For $K$ robots the bounding box for each horizon step $n$ is divided into $K$ equally sized smaller bounding boxes around $\+x_t^B$ (virtual leader). Each smaller bounding box can be represented as a linear constraint to guarantee inter-base collision avoidance. To ensure manipulation feasibility and avoid singular configurations, the dimensions of the regional constraint bounding box are geometrically constrained by the link lengths of the manipulator, predefined height $h_P$ of the payload and the dimensions of the cuboidal payload.

\subsubsection{Inter-Manipulator Collision Avoidance}
\begin{figure}
	\begin{subfigure}{0.49\columnwidth}
		\centering
		\includegraphics[scale=0.58]{manipulator_schem.pdf}
		\caption{}
		\label{f:DHparams}
	\end{subfigure}
	\begin{subfigure}{0.49\columnwidth}
		\centering
		\includegraphics[scale=0.36]{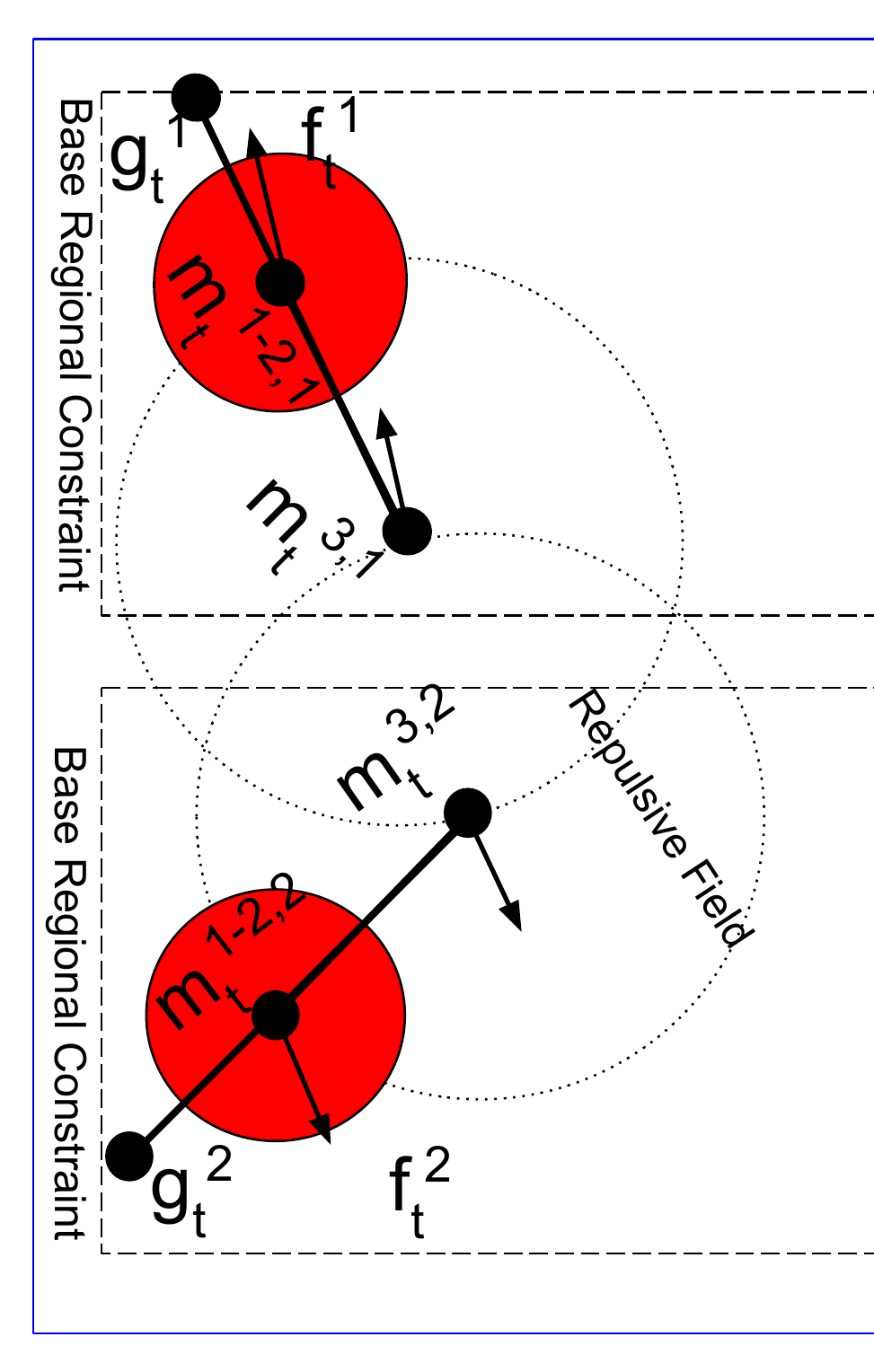}
		\caption{}
		\label{f:avoidance}		
	\end{subfigure}
	\caption{(a) Configuration of the 6-dof manipulator used in this work. $m_t^{1,k}, m_t^{2,k}$ are coincidental, $m_t^{4-6,k}$ form the coaxial 3-axis wrist (b) Inter-manipulator collision avoidance using artificial repulsive potential fields.}
	\label{f:manipulator}	
\end{figure}

If a mobile base nears the edges of its regional-constraints, the manipulator elbow joints $m_t^{3,k}$ could collide with neighboring manipulators. To avoid this scenario, we communicate the position of $m_t^{3,k}$ with other robots $j\neq k$. We then associate a repulsive field $F^i(d_{elbow})$ (see (\ref{eq:cotforce})) as a function of inter-robot elbow distances $d_{elbow}$.  Since the motion of base directly affects $\theta_t^{1,k}$ and $\mathbf{m}_t^{2-3,k}$, the repulsive field acting on each elbow joint directly translates to the mobile base. We incorporate this field as an external control input $\mathbf{f}_t^k$ for mobile base $k$ in MPC dynamics. The region of influence of this repulsive field is equal to $max(\|m_t^{2,k}-m_t^{3,k}\|_2,~\|m_t^{3,k}-m_t^{4,k}\|_2)$. This is highlighted using a dotted circle in Fig. \ref{f:avoidance}. 

\begin{figure*}[h]
	\begin{subfigure}{0.49\linewidth}
		\centering
		\includegraphics[width=\linewidth,trim={6cm 4.5cm 5cm 3cm},clip]{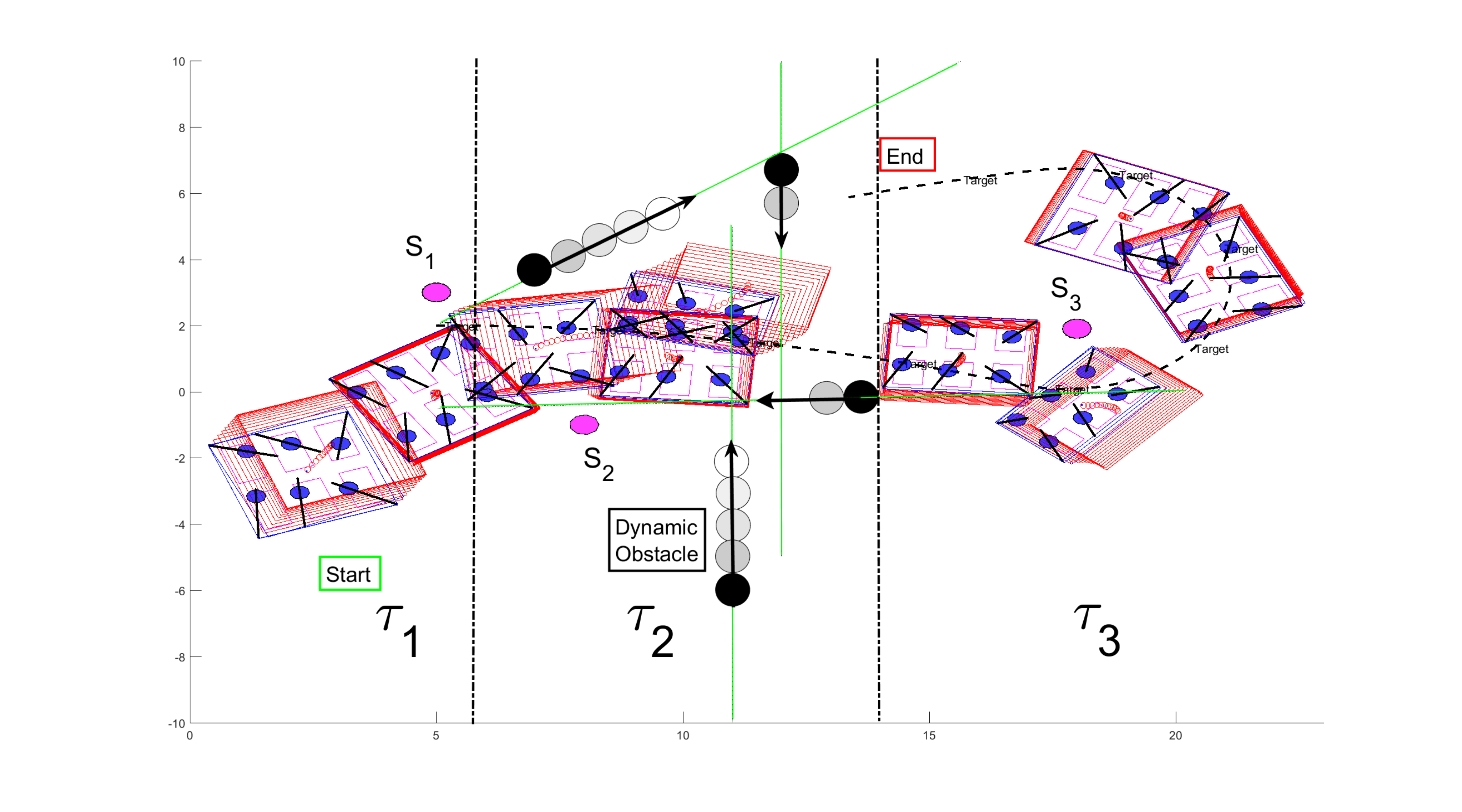}
		\caption{}
		\label{f:top}
	\end{subfigure}
	\begin{subfigure}{0.49\linewidth}
		\centering
		\includegraphics[scale=0.31,trim={10cm 0cm 7cm 11.5cm},clip]{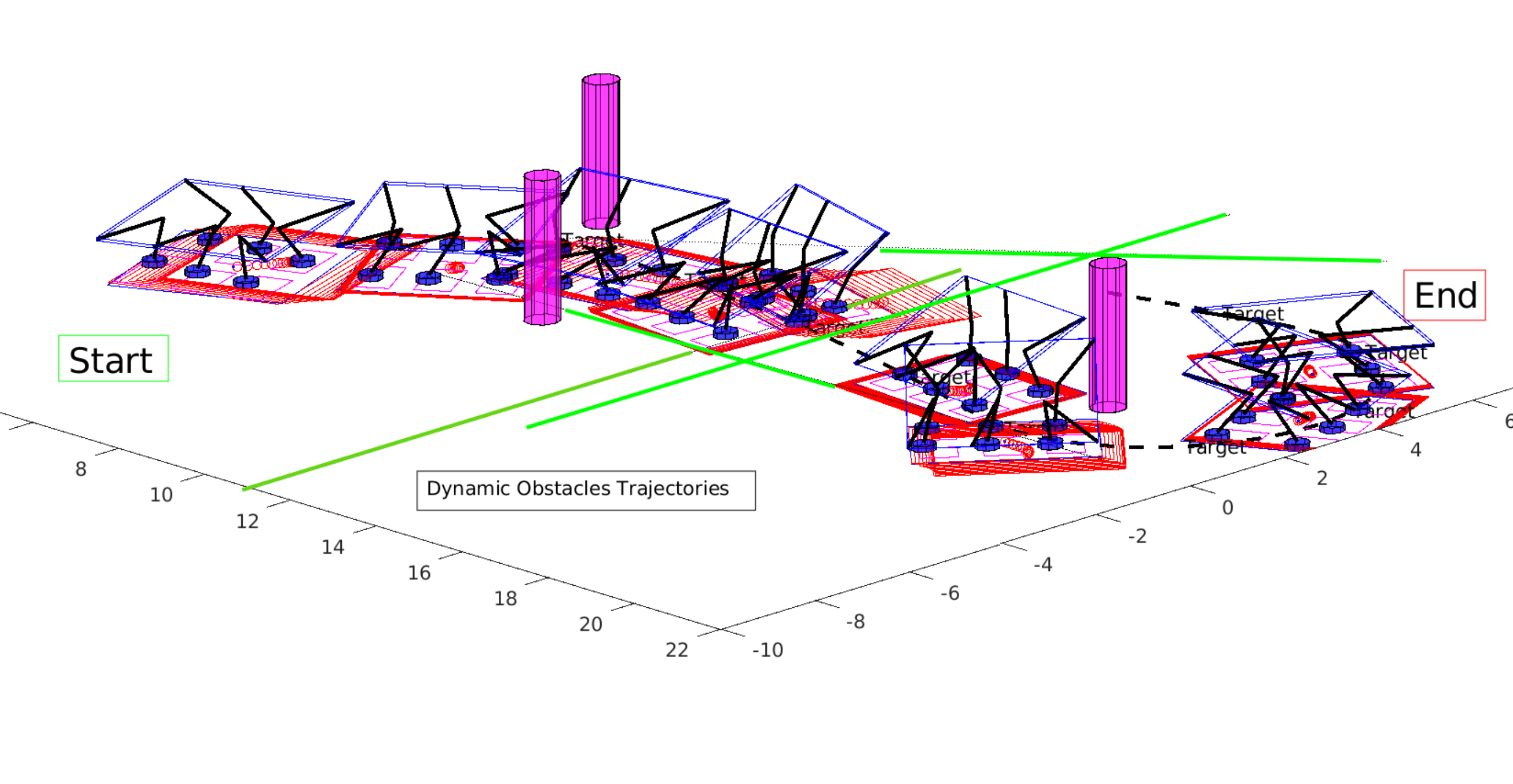}
		\caption{}
		\label{f:iso}		
	\end{subfigure}
	\caption{(a) Top view and (b) Isometric view of the system navigating through tight spaces. The DVBs over the planning horizon are visualized in red. The robots are represented as blue cylindrical mobile bases with mounted manipulators (in black). The payload is visualized in blue. }
	\label{f:topiso}	
\end{figure*} 

\subsubsection{Cost}
The goal of formation motion planning is not only to avoid collisions but also to ensure that the mobile bases minimize energy consumption and compute feasible trajectories over a prediction horizon $H_r \leq H$.
We therefore minimize the robot control input for a horizon and guide the robots towards the center of their respective regional constraints over $H_r$. The cost is defined as,
\begin{align} \label{JRMPC}
&J_{DMPC} = \sum_{n=0}^{H_r} \ao^k_t(n) \boldmath{\Omega}_{\ao} \ao^k_t(n)^\top + \nonumber \\ 
& (\x^k_t(n+1)-\x^k_{ref}(n+1)) \boldmath{\Omega}_\x (\x^k_t(n+1)-\x^k_{ref}(n+1))^\top
\end{align}
where, $\ao^k_t(n)\in \mathbb{R}^2$ is the linear translational velocity control input for robot $k$ at control horizon step $n$, $\x^k_{ref}(n+1)$ is the trajectory of the center of the regional constraint of $k$ over the control horizon and $\boldmath{\Omega}_{\ao}, ~\boldmath{\Omega}_\x$  are diagonal positive semi-definite weight matrices.
\subsubsection{Formation Trajectory Optimization}
The optimal trajectory ($\x^{k*}_t\leftarrow \text{state}, ~ \ao^{k*}_t\leftarrow \text{control}$ ) for each robot $k$ is computed using a decentralized MPC given by,
\begin{equation} \label{RMPC}
\x^{k*}_t(1)\dots \x^{k*}_t(H_r+1), ~ \ao^{k*}_t(0) \dots \ao^{k*}_t(H_r) = \argmin{\ao^k_t(0) \dots \ao^k_t(H_r)}  J_{DMPC} \nonumber 
\end{equation} 
subject to, 
\begin{align}
&  \x_t^{k}(n+1)^\top  =  \mathbf{A} \x_t^{k}(n)^\top  + \mathbf{B}(\ao_t^{k}(n)+\+f_t^k)^\top \label{rdynamics}\\
& \mathbf{Reg}_t^k(n+1) \x^k_t(n+1)^\top \leq \mathbf{b}_t^k(n+1) \label{Rregcon} \\
& \ao^k_\mathrm{min} \leq ~ \ao_t^{k}(n) \leq \ao_\mathrm{max}, \\
& \x^k_\mathrm{min} \leq ~ \x_t^{k}(n+1) \leq ~ \x^k_\mathrm{max}.
\end{align}
Here, (\ref{rdynamics}) is the linear translational dynamics for robot $k$. $\+f_t^k$ embeds the inter-manipulator obstacle avoidance constraint and is constant over the horizon $H_r$. Input $\+f_t^k$ is clamped to $0.5 \|\ao^k_{max}\|$ and $\ao^k_{max}\geq \ao^B_{max}$. The  DMPC is constrained to choose the $\ao_t^k$ in the range of $0.5 \ao^k_\mathrm{min} \leq ~  \ao_t^{k}(n) \leq 0.5 \ao^k_\mathrm{max}$ for  $\|\+f_t^k\|>0$. This ensures that $\+f_t^k$ does not force the robot out of its defined regional constraint. However, if $\|\+f_t^k\|=0$ at $t$, then the MPC controls the full range of $\ao_t^k$.  Eq. (\ref{Rregcon}) is a linear regional constraint for robot $k$ over the control horizon where, $\mathbf{Reg}^k_t(n+1) \in \mathbb{R}^{4\times2}$ and $\mathbf{b}^k_t(n+1)\in \mathbb{R}^{4}$. $\ao_t^{k*}(0)$ is used as control input for mobile base $k$ at time $t$. 
A feasible inverse kinematics solution can be computed using $\+g_t^k$ and $\x^{k*}_t(1)$ and the equations mentioned in Sec. \ref{sec:invkin}. 


\section{Results and Discussion} \label{sec:results}
In this section, we detail the results of our proposed motion planning algorithm. The DVB MPC and the decentralized formation MPC are both numerically solved using the operator splitting quadratic program (OSQP) solver \cite{osqp}. 
The payload motion planning is solved as a sequential quadratic program. 
The proposed algorithm is implemented in Matlab and runs on an Intel i7-7700HQ CPU. 
The algorithm is validated using multiple simulation environments varying in, (i) the number of static obstacles and dynamic obstacles, (ii)  the number of robots, (iii) the dimensions of payloads, and, (iv) trajectories of a moving target. These results are best viewed in the video {\footnotesize \url{https://youtu.be/9EKj7RwRs_4}}.
For the sake of analysis and brevity, we investigate and discuss one of these environments in detail. 

We consider an environment having three closely spaced, static obstacles and four randomly moving dynamic obstacles. A target $T$ navigates at approximately $0.4~ms^{-1}$ to provide high-level motion goals to the system. Six mobile manipulators are tasked with transporting a payload of dimensions $3~m\times3~m\times0.1~m$ through this environment while tracking $T$. 
Fig. \ref{f:topiso} showcases top and isometric views as the system navigates through tight spaces, using the proposed hierarchical motion planning algorithm.  The dashed black line is the trajectory followed by $T$. The three static obstacles are visualized as pink cylinders, and the four dynamic obstacles are visualized as black cylinders. Green lines highlight the trajectories of dynamic obstacles. The motion of the system is best visualized in the following video {\footnotesize \url{https://youtu.be/9EKj7RwRs_4}}. Dynamic obstacles are adversarial and hinder the motion of the system as observed in Fig. \ref{f:topiso} and the video.

\begin{figure}
	\centering
	\includegraphics[width=\columnwidth,trim={0.5cm 0.5cm 1cm 0.5cm}]{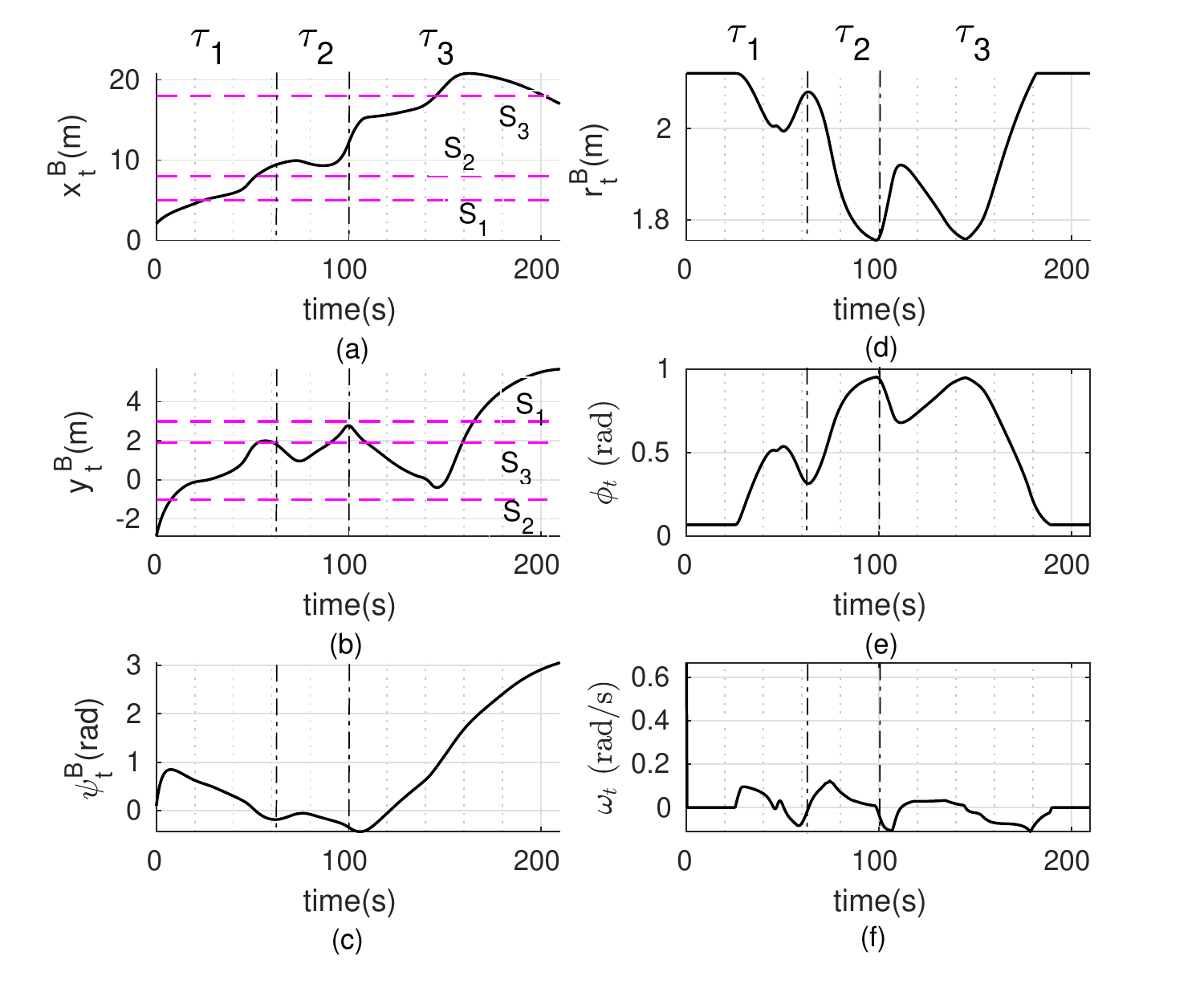}
	\caption{(a-d) DVB trajectory and scale. (e-f) payload roll and roll angular velocity.}
	\vspace{-0.5cm}
	\label{f:payload}
\end{figure}

\textit{Deformable Virtual Bounding Box: } The state and control limits are $[-30,-30] \leq~ \x_t^B~ \leq [30,30]$ in $m$ and $[-2,-2]~ \leq \ao_t^B ~\leq [2,2]$ in $ms^{-1}$ respectively. The control horizon $H$ has an effect on real time performance. For $H=12$ time steps and $\Delta t=0.1~s$, the average execution time $t_{avg}=0.17~s$. $d_{max} = d_{min}+1.8~m$ defines $\+f_t^B$'s region of influence with an $F_{max}=2.5~ms^{-1}$. The limits on DVB scale $1.3~m \leq r_t^B \leq 2.12~m$ are defined using the projection of payload for $\phi_t=0$ and $\phi_t=\phi_{max}$ respectively. In Fig. \ref{f:topiso} the red rectangles shows the motion of the DVB for $H$ time steps. 
The payload (blue rectangle) orients itself to lie within the defined workspace. Each mobile base (blue cylinders) operates within its regional constraints (small pink rectangles) and each manipulator (in black) has a collision-free kinematically feasible configuration. 
Fig. \ref{f:payload}(a-d) showcases the DVB pose $\x_t^B = [x_t^B~y_t^B~ \psi_t^B]$ and  scale $r_t^B$ over the experiment duration. 
Notice that Fig. \ref{f:top} and Fig. \ref{f:payload} are divided into three smaller timelines $\tau_1,~\tau_2,~\tau_3$. 
In $\tau_1$, the DVB deforms from $r_t^B=2.12~m$ to $r_t^B=2~m$ to avoid a dynamic obstacle and subsequently expands for about $10~s$. 
In $\tau_2$, the DVB navigates through a narrow gap between static obstacles $S_1,~S_2$ causing a decrease in $r_t^B$. Next, it encounters multiple dynamic obstacles causing a sharp decrease to $r_t^B=1.75~m$, as observed in Fig. \ref{f:payload}(d). 
In $\tau_3$, $r_t^B$ increases to $1.92~m$ until the DVB performs a sharp left turn around $S_3$ to keep track of $T$. Finally, $r_t^B$ gradually increases $2.12~m$ as obstacles become sparse.
The repulsive fields $\+f_{sta}$, $\+f_{dyn}$ aid in maintaining a distance of at least $r_t^B$ w.r.t static and dynamic obstacles throughout $\tau_1,~\tau_2$ and $~\tau_3$. In obstacle-free regions, $f_t^e$ restores $r_t^B$ to $r_{max}$. In Fig.\ref{f:payload}(a,b) the three static obstacle positions $S_1,S_2,S_3$ are overlaid (in pink). Observe that $x_t^B$ and $y_t^B$ simultaneously do not intersect the same pink line $(S_i)$ at the same exact time instant. This validates static obstacle avoidance. 
\\
\textit{Payload Motion Planning: } The limits on motion planning are $ 0~ \leq~ \phi_t ~\leq \frac{2 \pi}{5} $ in $rad$ and $-1 \leq ~\omega_t(n) ~\leq 1 $ in $rads^{-1}$. For a control horizon $H_p=5$ and $\Delta t = 0.1~s$, $t_{avg}=0.05~s$ was observed. The payload's roll $\phi_t$ varies in accordance with $r_t^B$ using non-linear optimization of Sec. \ref{sec:pmp}. Fig. \ref{f:payload}(e,f) plots the variation of $\phi_t$ and $\omega_t$. We observe that the payload rolls in accordance with $r_t^B$ and varies smoothly along the trajectory thereby validating the proposed optimization. \\
\begin{figure}
	\begin{subfigure}{0.5\textwidth}
		\centering
		\includegraphics[width=\columnwidth,trim={0.5cm 0.5cm 0cm 0.0cm}]{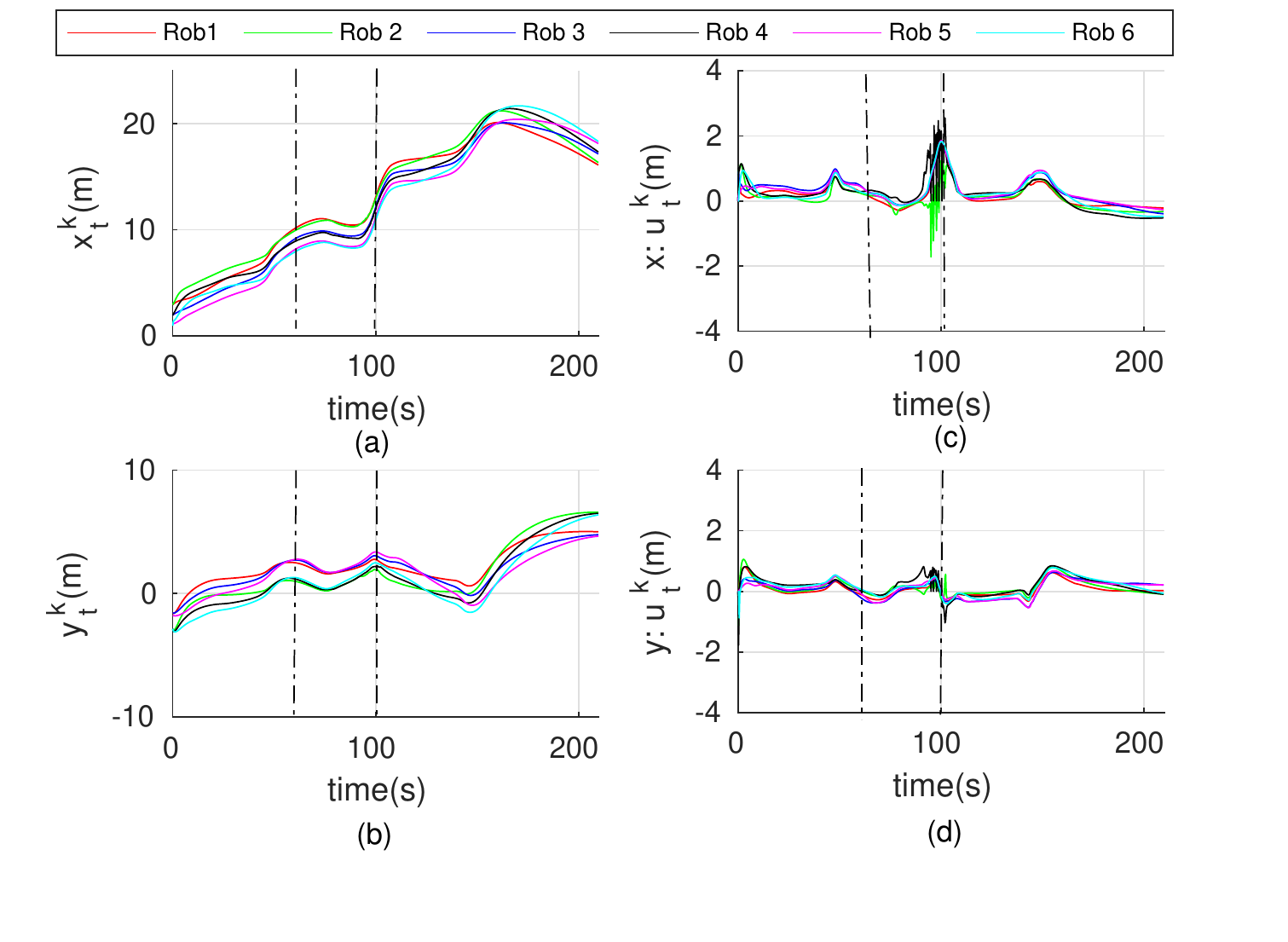}
	\end{subfigure}\vspace{-0.3cm}
	\begin{subfigure}{0.5\textwidth}
		\centering
		\includegraphics[width=\columnwidth,trim={0.5cm 0.5cm 0cm 0.0cm}]{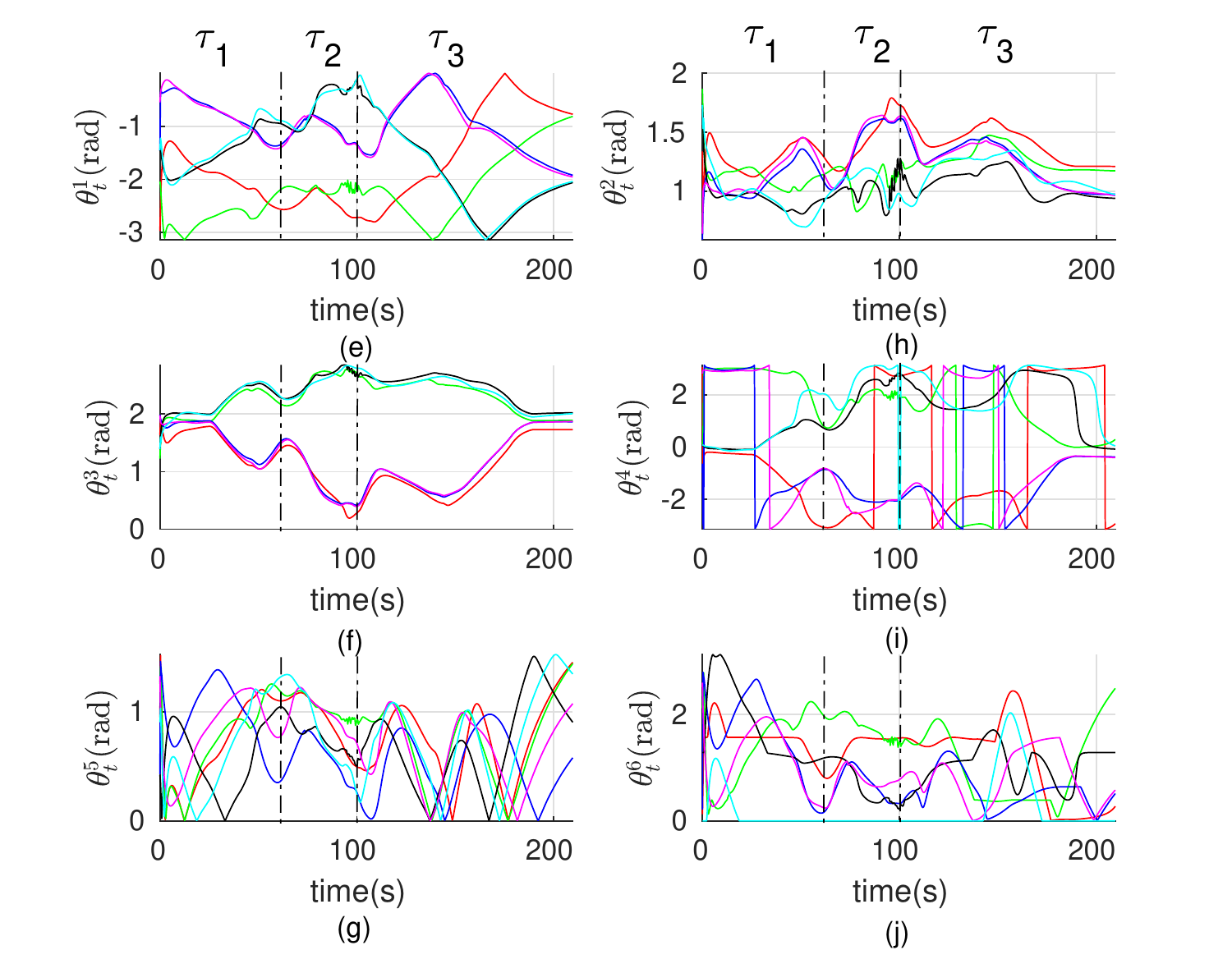}
	\end{subfigure}
	\caption{(a-d) Mobile base position and control input. (e-j) Manipulator joint angles.}
	\label{f:dmpc}
	\vspace{-0.4cm}
\end{figure} 
\textit{Decentralized Formation Motion Planning: } Each mobile base is a cylinder of radius $0.2~m$ and height $0.2~m$. The control limits of the mobile base are $-4 \leq~ \ao_t^k~\leq 4$ in $ms^-1$. 
A $t_{avg} = 0.08~s$ was observed for $H_r=5$. 
In this experiment, $m_t^{1,k}=m_t^{2,k}$ and $m_t^{4-6,k}=g_t^k$ (co-axial wrist). The link lengths are, $\|m_t^{2,k}-m_t^{3,k}\|_2 = 1.316~m$, $\|m_t^{3,k}-m_t^{4,k}\|_2 = 1.484~m$. The results of decentralized robot motion planning are showcased in Fig. \ref{f:dmpc}(a-d). The trajectories $\x_t^k=[x_t^k~y_t^k]$ are observed to be centered around $\x_t^B$ (virtual formation leader) and vary smoothly over time. The control inputs $\ao_t^k$ vary gradually over the trajectory and lie within the limits defined by the MPC. The high frequency changes in $\ao_t^k$ around $100~s$ ($\tau_2$), in Fig. \ref{f:dmpc}(c,d), can be attributed to low inter-manipulator spacing due to low $r_t^B$ (see Fig. \ref{f:payload}(d)), leading to high $\+f_t^k$. Fig. \ref{f:dmpc}(e-j) showcase the variation in the joint angles of manipulators over time. Notice that the variation is smooth and the velocities of the joints lie in the range of $[-1~rads^{-1},1~rads^{-1}]$. Note that the variations in $\theta_t^{4}$ are just angle wraparounds. These plots validate the efficacy of the proposed decentralized formation controller.

\vspace{-0.2em}
\section{Conclusions and Future Work} \label{sec:conclusions}
\vspace{-0.2em}
In this paper, a novel kinematic motion planning algorithm for cooperative mobile non-planar payload manipulation in dynamic environments is presented. Three constrained optimization problems are formulated to handle key challenges namely,  (1) computationally scalable goal-directed non-planar manipulation, (2) environmental obstacle avoidance, and, (3) inter-robot obstacle avoidance. In future, we plan to physically validate the algorithm on real robots.

\section*{Acknowledgments}
We thank Eric Price for his valuable advice during the course of this work.

\bibliographystyle{IEEEtran}
\bibliography{paper}

\end{document}